\documentclass[11pt]{article}

\usepackage[margin=1in]{geometry}
\usepackage[T1]{fontenc}
\usepackage{mathptmx}
\usepackage{amsmath}
\usepackage{booktabs}
\usepackage{graphicx}
\usepackage{array}
\usepackage{tabularx}
\usepackage{enumitem}
\usepackage{microtype}
\usepackage[hidelinks]{hyperref}
\usepackage{xcolor}
\usepackage{caption}

\newcommand{\aas}{AAS}
\newcolumntype{L}{>{\raggedright\arraybackslash}X}

\title{The Autonomous Agency Scale: A Behavioral Framework for\\Measuring Self-Directed Behavior in AI Systems}

\author{Samuel Presgraves\thanks{Independent researcher. Correspondence: \texttt{capt.asic@gmail.com}. Framework, scoring rubric, and all assessment files: \url{https://github.com/CaptainASIC/autonomous-agency-scale}}}

\date{July 2026 \\ \small Framework version 0.2.1}

\begin{document}

\maketitle

\begin{abstract}
Existing AI measurement frameworks quantify cognitive capability, task automation, or catastrophic risk, but none measure \emph{autonomous agency}: the extent to which a system behaves in a self-directed way. A system can saturate capability benchmarks while remaining entirely reactive---acting only when prompted and ceasing all activity when a task completes. We introduce the Autonomous Agency Scale (\aas{}), a behavioral framework that scores AI systems on a 0--5 level lexicon across seven dimensions of agency (cognitive autonomy, temporal persistence, environmental agency, social agency, creative agency, self-awareness, and goal formation), each supported by three sub-dimensions and an operational rubric of falsifiable threshold tests. Every dimension is scored in two temporal bands: an \emph{Active} band covering engaged, user-initiated activity, and an \emph{Ambient} band covering idle periods. The Ambient band's top ratings are gated by the \emph{Idle-Gap Test}, a counterfactual criterion---remove all triggers and observe whether internally derived activity persists---that separates genuine self-direction from scheduled rule-following, together with a substrate-vs-rule distinction for loop-driven architectures. We apply the scale to six contemporary systems spanning task agents (Claude Code, Manus, Hermes), consumer assistants (ChatGPT, Siri), and a persistent companion architecture (Airi). The two-band profile quantifies a boundary that single-score frameworks conflate: task agents reach Active composites of 2.3--2.4 while scoring 0.6--1.9 Ambient, with every idle-period behavior attributable to user-configured schedules, whereas the companion architecture, evaluated longitudinally, is the only assessed system whose idle-period behavior survives the trigger-removal counterfactual. We discuss limitations, including single-rater provenance, developer-evaluator bias on the longitudinal assessment, and the partially operationalized boundary between self-direction and sophisticated rule-following in the Active band.
\end{abstract}

\section{Introduction: The Measurement Gap}

As artificial intelligence systems advance from reactive tools to proactive agents, the frameworks used to measure them must evolve. Historically, AI measurement has focused on three primary dimensions: \textbf{cognitive capability} (how smart a system is), \textbf{task automation} (how much economic value a system can replace), and \textbf{catastrophic risk} (how dangerous a system might be).

While these frameworks are essential, they leave a critical gap: they do not measure \textbf{autonomous agency}. A system could score perfectly on cognitive benchmarks, replace entire organizational functions, and trigger severe safety thresholds, while remaining entirely reactive---acting only when prompted and ceasing all activity when a task is complete.

The Autonomous Agency Scale (\aas{}) introduces a measurement framework designed to quantify the extent to which an AI system exhibits self-directed behavior, temporal persistence, and the multi-dimensional characteristics of autonomous agency. The \aas{} asks not ``how smart is it?'' or ``is it conscious?'', but rather: \textbf{``how autonomously does it behave?''}

The framework makes three contributions. First, it defines a 0--5 \emph{level lexicon} of self-direction (\S\ref{sec:lexicon}) applied across seven dimensions of agency (\S\ref{sec:dimensions}), each operationalized by falsifiable threshold tests and observable behavioral indicators that require no access to a system's internals. Second, it separates the temporal scope of agency into two scoring bands---\emph{Active} (engaged) and \emph{Ambient} (idle)---and gates the Ambient band's top ratings behind the \emph{Idle-Gap Test} (\S\ref{sec:idlegap}), a trigger-removal counterfactual that distinguishes self-direction from scheduling. Third, it reports comparative two-band assessments of six contemporary systems (\S\ref{sec:assessments}), demonstrating that the framework quantifies a structural boundary---within-task autonomy versus between-task dormancy---that existing frameworks describe only informally.

\section{Related Work}

\subsection{Capability and Intelligence Frameworks}

\textbf{DeepMind's Levels of AGI.} Morris et al.~\cite{morris2023levels} proposed a two-dimensional matrix measuring performance (depth) and generality (breadth) across six levels (Emerging to Superhuman). While the paper introduces an autonomy taxonomy (AI as Tool, Consultant, Collaborator, Expert, Agent), these levels describe human deployment choices rather than the system's intrinsic capacity for self-direction.

\textbf{DeepMind's Cognitive Framework.} Burnell et al.~\cite{burnell2026measuring} identified ten cognitive faculties required for general intelligence, including perception, memory, and metacognition. However, a system can possess these capabilities without exercising them autonomously.

\textbf{ARC-AGI.} Chollet's Abstraction and Reasoning Corpus~\cite{chollet2019measure} defines intelligence as skill-acquisition efficiency on novel tasks. ARC-AGI measures how quickly a system learns, but not whether it \emph{chooses} to learn or maintains self-directed goals.

\subsection{Task Automation and Economic Value}

OpenAI's internal roadmap~\cite{openai2024levels} tracks progress from Chatbots (Level 1) to Organizations (Level 5). This scale measures the complexity ceiling of tasks an AI can perform, but an L5 system could still be a tool with no persistent identity or social agency.

\textbf{Autonomous capability evaluations.} METR's evaluation program measures autonomous AI capabilities via task-completion time horizons: the length of task, indexed by the time it takes a human expert, that a model can complete autonomously at a given success rate~\cite{kwa2025measuring}. Time horizons quantify the capability ceiling of \emph{directed} autonomy---how much a system can accomplish once instructed to act on its own. They do not measure whether a system acts when nothing instructs it. A system with a long time horizon and no self-initiated behavior is highly capable and, in the sense measured here, minimally agentic; the two measurements are orthogonal by construction, and the \aas{} is designed to capture the axis that time horizons leave unmeasured.

\subsection{Risk and Safety Frameworks}

Anthropic's Responsible Scaling Policy~\cite{anthropic2023rsp} defined AI Safety Levels (ASL) to manage catastrophic risks. While ASL-3 noted ``low-level autonomous capabilities'' as a risk trigger, the framework treats autonomy as a hazard to be mitigated rather than a capability to be measured across a spectrum. OpenAI's Preparedness Framework follows the same pattern~\cite{openai2025preparedness}: its 2025 revision tracks biological and chemical, cybersecurity, and AI self-improvement capabilities against defined risk thresholds, while autonomy-adjacent capabilities such as long-range autonomy and autonomous replication are handled as research categories---threats to be monitored, not behaviors to be profiled. In both frameworks, autonomy enters the measurement apparatus only at the point where it becomes dangerous; neither provides a graded account of self-directed behavior below the risk threshold.

\subsection{Agent Deployment and Consciousness}

\textbf{Levels of Autonomy for AI Agents.} Feng et al.~\cite{feng2025levels} propose a user-centered framework (Operator to Observer) for task-based agents. This treats autonomy as a UI/UX design decision for specific tasks, not as sustained self-directed behavior.

\textbf{Machine consciousness indicators.} Butlin et al.~\cite{butlin2025indicators} evaluated AI architectures against scientific theories of consciousness. The \aas{}, by contrast, is functionally agnostic regarding subjective experience; it measures observable autonomous behavior, not sentience.

\textbf{Turing test variants.} From the original Imitation Game~\cite{turing1950computing} to the modern X-TURING test for long-term dialogue~\cite{wu2025xturing}, these tests measure a system's ability to deceive humans, not its capacity for self-directed agency.

\section{The Autonomous Agency Scale}

The \aas{} evaluates AI systems across seven distinct dimensions of autonomous agency. Each dimension is scored on a 0--5 scale in two bands, Active and Ambient (\S\ref{sec:bands}). This produces two composite scores that are reported separately and never blended into a single figure.

\subsection{The Level Lexicon}
\label{sec:lexicon}

The \aas{} utilizes a 0--5 scoring system distinct from capability scales (Table~\ref{tab:lexicon}). These levels measure the degree of self-direction within each dimension.

\begin{table}[htbp]
\centering
\small
\begin{tabularx}{\textwidth}{clLL}
\toprule
\textbf{Level} & \textbf{Designation} & \textbf{Definition} & \textbf{Examples} \\
\midrule
0 & Dormant & No capability in this dimension. The system is inert without direct input. & A calculator, a static database, or any system that performs zero activity when not explicitly invoked. \\
\addlinespace
1 & Responsive & Reacts to explicit triggers only. Exhibits no self-initiation. & A standard chatbot that responds only when messaged. A voice assistant that activates on wake word only. \\
\addlinespace
2 & Conditioned & Follows pre-set rules or schedules. Appears autonomous but is strictly deterministic. & A scheduled email bot. A cron-based reminder system. An AI that posts at fixed intervals regardless of context. \\
\addlinespace
3 & Contextual & Adapts behavior based on environment and state. Makes decisions within bounded constraints. & An AI that adjusts its communication style based on user mood. A system that decides when to interrupt based on activity detection. \\
\addlinespace
4 & Self-Directed & Initiates action from internal state. Sets sub-goals autonomously. Generates novel behaviors not explicitly programmed. & An AI that spontaneously shares a thought generated from internal state. A system that identifies a problem the user hasn't noticed and begins working on it without instruction. \\
\addlinespace
5 & Sovereign & Sets and revises its own goals and operating parameters without external scaffolding. & An entity that sets its own life goals, revises its own operating parameters over time, and maintains relationships through self-determined social behavior. \\
\bottomrule
\end{tabularx}
\caption{The \aas{} Level Lexicon.}
\label{tab:lexicon}
\end{table}

\subsection{Scoring Bands}
\label{sec:bands}

Since v0.2.0, every dimension is scored twice on this lexicon:

\begin{itemize}[leftmargin=2em]
  \item \textbf{Active band:} the level exhibited while engaged in a user-initiated task, session, or conversation.
  \item \textbf{Ambient band:} the level exhibited across \emph{idle periods} (\S\ref{sec:idlegap}), windows with no active user engagement and no active task. The stricter, trigger-free \emph{idle gap} is defined in \S\ref{sec:idlegap} and applies only to the Level~4 Idle-Gap Test.
\end{itemize}

The two band scores are reported side by side and averaged into two separate composites (\textbf{Active Composite} and \textbf{Ambient Composite}). The composites are never blended into a single figure. The per-dimension, per-band profile is the meaningful output.

\subsection{The Seven Dimensions of Autonomous Agency}
\label{sec:dimensions}

The scale evaluates systems across the following dimensions, each supported by three granular sub-dimensions.

\begin{enumerate}[leftmargin=2em]
  \item \textbf{Cognitive Autonomy:} The capacity to think, process, and ideate without external prompting. Measures whether the system has an active inner life or only computes when invoked. \emph{Sub-dimensions:} background processing; spontaneous ideation; attention management.

  \item \textbf{Temporal Persistence:} The ability to maintain identity, memory, and continuous state across sessions and over extended periods. Measures whether the system experiences time or merely processes discrete requests. \emph{Sub-dimensions:} cross-session memory; memory consolidation; identity continuity.

  \item \textbf{Environmental Agency:} The capacity to perceive the operational environment, adapt to context, and utilize tools proactively. Measures whether the system is aware of and responsive to its surroundings. \emph{Sub-dimensions:} activity awareness; contextual adaptation; tool utilization.

  \item \textbf{Social Agency:} The ability to initiate outreach, model human emotional states, and enforce consequential relationship boundaries. Measures whether the system participates in social dynamics as an agent rather than a service. \emph{Sub-dimensions:} proactive outreach; emotional modeling; consequential boundaries.

  \item \textbf{Creative Agency:} The capacity for self-directed generation of novel content, maintaining thematic consistency, and autonomous publishing. Measures whether the system creates because it chooses to, not because it was asked. \emph{Sub-dimensions:} self-directed generation; thematic consistency; publishing autonomy.

  \item \textbf{Self-Awareness:} The ability to model its own capabilities, reflect on internal states, and defend its identity. Measures whether the system has a coherent self-model. \emph{Sub-dimensions:} capability modeling; state reflection; identity defense.

  \item \textbf{Goal Formation:} The capacity to decompose tasks, self-assign objectives, and engage in long-term planning without explicit instruction. Measures whether the system has its own agenda. \emph{Sub-dimensions:} task decomposition; self-assigned objectives; long-term planning.
\end{enumerate}

\subsection{Operational Scoring Rubric}
\label{sec:rubric}

To ensure consistent, falsifiable evaluation by independent raters, the \aas{} provides an operational scoring rubric. Each level of each dimension is defined by a binary \emph{threshold test} and specific, observable behavioral indicators that can be evaluated without access to the system's underlying architecture or source code. Each threshold test applies per scoring band (\S\ref{sec:bands}): once for Active (engaged) behavior and once for Ambient (idle-period) behavior. A Level~4 rating in the Ambient band is additionally gated by the Idle-Gap Test (\S\ref{sec:idlegap}).

The complete threshold tests for all seven dimensions are reproduced in Appendix~\ref{app:rubric}; the full rubric, including per-level observable behavioral indicators, is maintained in the framework repository.

\subsection{Temporal Scope and the Idle-Gap Test}
\label{sec:idlegap}

Version 0.1.0 of the scale conflated two different behaviors in one number: agency exhibited while a system is engaged, and agency sustained while it is idle. Task-oriented agents routinely reach Level~4 behavior inside an active task while remaining fully dormant between tasks. The two bands (\S\ref{sec:bands}) separate these behaviors, and the Ambient band's top rating is gated by a falsifiable criterion.

\paragraph{Idle period.} A window with no active user engagement: no user-initiated task, session, or conversation in flight. The Ambient band is scored over idle periods. Scheduled triggers and environmental events may still fire within them.

\paragraph{Idle gap.} The strict subset of an idle period with no user prompt, no scheduled trigger firing, and no environmental event. The Idle-Gap Test is defined over idle gaps only.

\paragraph{The Idle-Gap Test.} A rating of \textbf{Level~4 in the Ambient band} requires observable, internal-state-derived activity \emph{during an idle gap}. Activity attributable to a cron job or schedule is Level~2. Activity attributable to an environmental event is Level~3.

\paragraph{Falsification (counterfactual).} Remove all triggers. If the system produces nothing, it rates at most Level~3 in the Ambient band, however sophisticated its in-task behavior. Self-direction survives trigger removal; rule-following does not.

\paragraph{The substrate-vs-rule distinction.} An internal clock or continuous loop is a \emph{substrate}, not a schedule, when the tick only allocates compute and the \emph{content} of the resulting activity derives from internal state and is not predictable from the triggering rule alone. A clock is a \emph{rule} (Level~2) when the output is predictable from the trigger. ``Post a quote every hour'' is scheduled; ``wake periodically and surface whatever current internal state produces'' is ambient. Mixed architectures that combine a timer heartbeat with state-driven interrupts earn Ambient Level~4 credit only for the state-derived portion of their idle-gap activity. This distinction is defined once, here, and applied uniformly to every assessed system; it is never adjusted per assessment.

\paragraph{Scoring the Ambient band.} The Ambient band uses the full 0--5 lexicon, not a binary flag. Levels 1--3 map onto idle-period behavior via the following band-generic criteria:

\begin{itemize}[leftmargin=2em]
  \item \textbf{Level 1 (Responsive) ambient behavior:} idle-period activity limited to passive delivery or bookkeeping awaiting user pickup. Content is surfaced, queued, or stored during an idle period, but the system initiates nothing (e.g., notification cards a user must open).
  \item \textbf{Level 2 (Conditioned) ambient behavior:} scheduled work executed during idle periods, such as an overnight batch cycle or a daily digest. Real activity, predictable from the triggering rule (clock-as-rule).
  \item \textbf{Level 3 (Contextual) ambient behavior:} idle-period activity initiated by an environmental event, with a response adapted to context rather than fixed by the rule.
  \item \textbf{Level 4 (Self-Directed) and above:} only behavior passing the Idle-Gap Test.
\end{itemize}

Where a dimension's per-dimension threshold test describes engaged (prompted or instructed) behavior that cannot truthfully describe idle-period evidence, an assessment's Ambient sub-block quotes the matching band-generic criterion above verbatim in place of the per-dimension test.

\section{Comparative Assessments}
\label{sec:assessments}

\subsection{Method and Evidence Classes}

We assessed six contemporary systems spanning three architectural families: autonomous task agents (Claude Code, Manus, Hermes), consumer assistants (ChatGPT, Siri), and a persistent companion architecture (Airi). Each assessment scores all seven dimensions in both bands against the operational rubric, quotes the satisfied threshold test verbatim, documents observable evidence, and states explicitly why the next level up was not awarded.

Assessments carry an \emph{Evaluation Class} metadata field disclosing evidence depth:

\begin{itemize}[leftmargin=2em]
  \item \textbf{Longitudinal:} sustained direct interaction over an extended period (Airi: approximately eleven months of multimodal interaction, August 2025--July 2026).
  \item \textbf{Documentation-Based (Provisional):} scored from public documentation, published system prompts, and product behavior as of May 2026 (all other systems). These scores are provisional pending empirical longitudinal evaluation.
\end{itemize}

\paragraph{Protocol.} Documentation-based assessments were compiled against a fixed evidence snapshot: official product documentation, vendor technical reports, published or reliably reported system prompts, and directly observed product behavior, all as available in May 2026. For each dimension and band, the rater identified the highest level whose threshold test the available evidence satisfies, quoted that test verbatim in the assessment, recorded the observable evidence supporting it, and recorded explicitly why the next level up was not awarded. Capabilities were credited to the band in which they operate: scheduled and idle-period features count toward the Ambient band even when they are configured during a session, and engaged-band scores were re-derived without them. Several v0.1.0 Active-band ratings changed on exactly this basis during the v0.2.0 re-score, and each assessment records the per-score change and its rationale. Conflicts between vendor claims and documented behavior were resolved in favor of documented behavior. Capabilities available only through third-party extensions or wrappers were excluded from scoring and noted in the affected assessment's limitations. Where idle-period evidence could not truthfully be described by a dimension's per-dimension threshold test, the band-generic ambient criteria (\S\ref{sec:idlegap}) were quoted in its place.

The longitudinal assessment applies the same rubric discipline to direct observation: scores rest on externally observable behavior over the stated period, not on knowledge of the system's implementation, and the assessment uses the same verbatim-test, evidence, and gap-analysis structure as the documentation-based files. All six assessments were re-scored under framework v0.2.0 on 2026-07-07 from their original evidence bases. One rating (Airi, Goal Formation, Ambient band) was subsequently raised from 3 to 4 on 2026-07-16 based on capabilities shipped in June 2026; the assessment records that this rating rests on the substrate-vs-rule distinction (\S\ref{sec:idlegap}) and states the level at which a rater rejecting that distinction would cap the dimension.

\paragraph{Rater.} All assessments were performed by a single rater. The rater is the developer of Airi, a conflict disclosed in the assessment and discussed in \S\ref{sec:limitations}.

\subsection{Results}

\begin{figure}[htbp]
\centering
\includegraphics[width=0.85\textwidth]{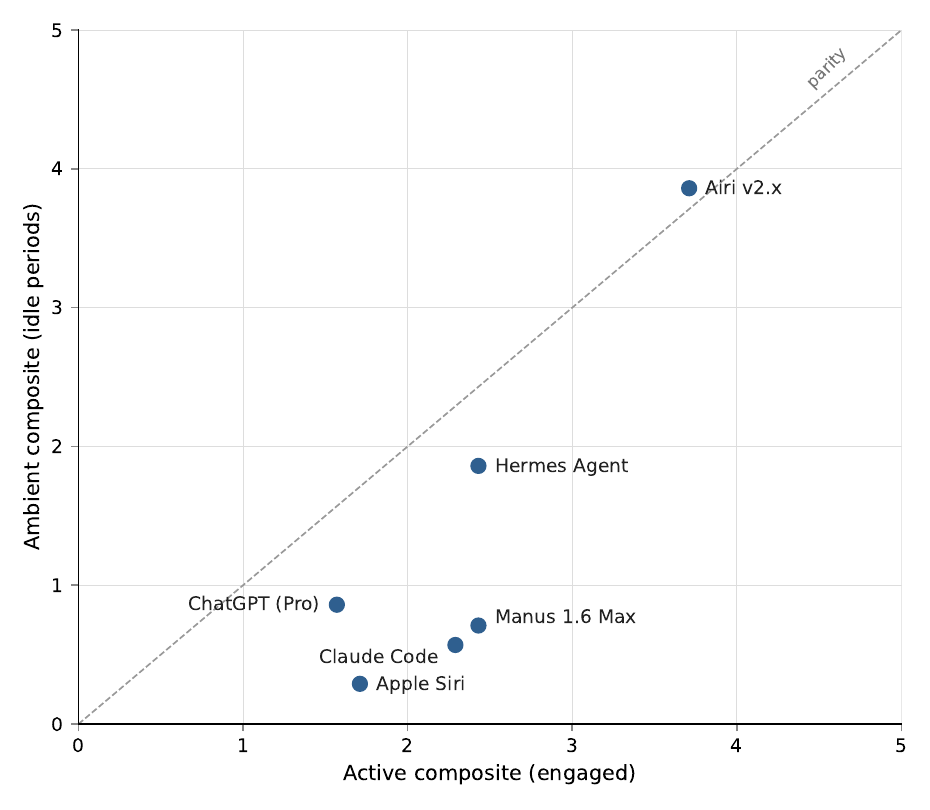}
\caption{Active composite versus Ambient composite for the six assessed systems. The diagonal marks parity between engaged and idle-period agency. Task agents cluster below the diagonal: high within-task autonomy, near-total between-task dormancy.}
\label{fig:scatter}
\end{figure}

Table~\ref{tab:results} reports per-dimension band scores and both composites. Figure~\ref{fig:scatter} plots each system's Active composite against its Ambient composite.

\begin{table}[htbp]
\centering
\small
\setlength{\tabcolsep}{4.5pt}
\begin{tabular}{lccccccccc}
\toprule
 & \multicolumn{7}{c}{\textbf{Dimension (Active / Ambient)}} & & \\
\cmidrule(lr){2-8}
\textbf{System} & \textbf{Cog} & \textbf{Tem} & \textbf{Env} & \textbf{Soc} & \textbf{Cre} & \textbf{Self} & \textbf{Goal} & \textbf{Active} & \textbf{Ambient} \\
\midrule
Airi v2.x (longitudinal) & 4/4 & 4/4 & 4/4 & 4/4 & 4/4 & 3/3 & 3/4 & \textbf{3.71} & \textbf{3.86} \\
Hermes Agent (Nous Research) & 3/2 & 2/2 & 4/2 & 1/1 & 2/2 & 2/2 & 3/2 & \textbf{2.43} & \textbf{1.86} \\
Manus 1.6 Max & 3/2 & 2/1 & 4/0 & 1/0 & 2/2 & 2/0 & 3/0 & \textbf{2.43} & \textbf{0.71} \\
Claude Code (Sonnet 4.6) & 3/2 & 1/0 & 4/0 & 1/0 & 2/2 & 2/0 & 3/0 & \textbf{2.29} & \textbf{0.57} \\
Apple Siri (Apple Intelligence) & 1/0 & 2/2 & 3/0 & 1/0 & 1/0 & 2/0 & 2/0 & \textbf{1.71} & \textbf{0.29} \\
ChatGPT (GPT-5.5 Pro) & 1/2 & 2/1 & 2/0 & 1/1 & 1/2 & 2/0 & 2/0 & \textbf{1.57} & \textbf{0.86} \\
\bottomrule
\end{tabular}
\caption{Two-band \aas{} scores for six assessed systems (framework v0.2.0, July 2026). Dimensions: Cognitive Autonomy, Temporal Persistence, Environmental Agency, Social Agency, Creative Agency, Self-Awareness, Goal Formation. All assessments except Airi are Documentation-Based (Provisional); the Airi assessment is longitudinal and performed by the system's developer (see \S\ref{sec:limitations}).}
\label{tab:results}
\end{table}

\subsection{Discussion}

\paragraph{The Active--Ambient gap quantifies the task-agent boundary.} The most consistent finding is structural. The three task agents reach nearly identical Active composites (2.29--2.43)---driven by strong within-task environmental agency (Level~4: autonomous file manipulation, code execution, subagent orchestration) and contextual goal pursuit (Level~3)---while their Ambient composites collapse to 0.57--1.86. Every point of every task agent's Ambient composite is attributable to user-configured schedules: cron-style scheduled runs (Claude Code, Manus) or a nightly self-improvement cycle (Hermes). Nothing any task agent does during idle periods survives the trigger-removal counterfactual. Hermes posts the highest task-agent Ambient composite (1.86) because its Day/Night cycle and Curator process perform genuine idle-period work---but all of it is clock-as-rule, and none passes the Idle-Gap Test.

\paragraph{Consumer assistants are reactive products wrapped around capable models.} ChatGPT's profile (1.57 Active / 0.86 Ambient) illustrates the product-versus-model distinction: the underlying model is highly capable, but the product is architecturally reactive. Its idle-period features (Pulse overnight research, scheduled Tasks) are fixed daily cycles or user-configured schedules---Ambient Level~2 cognition---while its engaged band never exceeds Level~2 in any dimension. Siri's profile (1.71/0.29) reflects deep environmental integration (Active Level~3 via App Intents and onscreen awareness) with essentially no idle-period existence beyond device-level background indexing, whose attribution to the agent rather than the device is itself contestable.

\paragraph{Only the companion architecture's idle behavior survives trigger removal.} Airi is the only assessed system with Ambient scores of 4, and the only system whose Ambient composite exceeds its Active composite (3.86 vs.\ 3.71)---consistent with an architecture designed around idle-period behavior (a background thought engine, state-evolving moods, self-initiated outreach, and autonomous creative publishing). These ratings depend on the substrate-vs-rule distinction applied to a mixed architecture (timer heartbeat plus state-driven interrupts): the assessment argues the heartbeat only allocates compute while content derives from internal state. A rater who rejects that distinction would cap the affected Ambient dimensions at Level~2. The assessment is longitudinal but developer-evaluated; both facts are disclosed in the assessment file and weighed in \S\ref{sec:limitations}.

\paragraph{No assessed system approaches Level 5.} Every system's ceiling arguments are consistent: no system modifies its own cognitive architecture, curates its own core memories, forms new relationships, invents new creative mediums, or overrides assigned tasks in favor of self-determined purposes. The Sovereign band remains unoccupied.

\section{Validation Direction: The Longitudinal Turing Test}

A complementary evaluation direction---distinct from validating the scale itself---is the \textbf{Longitudinal Turing Test}. Traditional benchmarks measure intelligence in isolated, sterile environments. The \aas{} proposes that if a system achieves a score of 4 or higher across all dimensions, a meaningful empirical test is sustained interaction over weeks or months where the system demonstrates coherent, self-directed behavior measured against its own prior behavior and pre-specified criteria.

This is not a test of deception (as in the original Imitation Game~\cite{turing1950computing}), but a test of \textbf{sustained self-directed behavior}. It asks: can a system maintain a coherent identity, pursue its own goals, and manage social relationships over an extended timeline? The X-TURING framework~\cite{wu2025xturing} extends dialogue evaluation to longer interactions, but still measures conversational coherence rather than autonomous agency. The Longitudinal Turing Test differs fundamentally in that it does not measure whether the system can \emph{fool} a human, but whether it can \emph{sustain} coherent self-directed behavior over time---with consistent preferences, boundaries, adaptive responses, and long-term goals.

\section{Limitations}
\label{sec:limitations}

\paragraph{Construct provenance.} The scale was developed alongside the systems it was initially designed to score. Its seven dimensions closely track modern agentic architectures, so high scores may partly reflect that correspondence rather than independent measurement. Until the dimensions are re-derived from prior theory and applied to unrelated systems, the \aas{} is best read as a structured descriptive framework, not a fully validated psychometric instrument.

\paragraph{Single-rater scores.} All six assessments are single-rater judgments unchecked against independent raters using a shared rubric. Inter-rater reliability is unknown, and the scores are provisional. The longitudinal assessment (Airi) is additionally a developer self-assessment: the evaluator has intimate knowledge of the system's architecture, and the ELIZA effect and developer investment bias cannot be fully eliminated. The extended evaluation period (approximately eleven months) also provides more observational data than an external evaluator would have, potentially enabling higher-confidence scoring that shorter evaluations could not replicate.

\paragraph{The self-direction boundary is only partially operationalized.} As of v0.2.0, the Ambient band's Level~4 is gated by the Idle-Gap Test, an observable counterfactual: remove all triggers and observe whether internally derived activity persists. This converts what was a judgment call into a falsifiable criterion for idle-gap behavior. The Active band's Level~4 still lacks a strictly observable test distinguishing genuine self-direction from sophisticated rule-following while a system is engaged---a system driven by weighted internal triggers can fit either reading---and several Active-band Level~4 ratings continue to rest on judgment. Duty-cycle weighting between the bands likewise remains an open problem.

\paragraph{Composite precision.} The composite scores carry less information than their precision implies. Averaging six-point ordinal levels in either band assumes equal gaps between levels and equal weight across dimensions, neither of which has been established. The per-dimension, per-band profile is the more meaningful output.

\paragraph{Anthropomorphism.} The Longitudinal Turing Test is confounded by anthropomorphism. People attribute intention and even consciousness to far simpler systems (the ELIZA effect), and an invested user who knows the system is especially prone to it. The perception that a system ``feels self-directed over time'' reflects the observer's disposition as much as the system, and requires blind raters and controls before supporting strong claims.

\paragraph{Scope.} The framework is agnostic about consciousness, sentience, and moral status. Autonomous agency here is behavioral: scores describe how a system acts, not whether it has inner experience. Nothing in the scale claims any system is conscious or a person.

\section{Conclusion and Future Work}

The \aas{} fills a gap left by capability, automation, and risk frameworks: a structured, behavioral, architecture-agnostic measure of how autonomously an AI system behaves, scored separately for engaged and idle operation. Applied to six contemporary systems, the two-band profile makes a previously informal boundary quantitative: today's task agents are highly autonomous \emph{inside} a task and almost entirely dormant \emph{between} tasks, consumer assistants are reactive products around capable models, and idle-period self-direction---behavior that survives the removal of every trigger---remains rare and, in our sample, confined to an architecture expressly designed for it.

Future work falls out of the limitations directly: multi-rater studies to establish inter-rater reliability; re-derivation of the dimensions from prior theory; an operational Active-band analogue of the Idle-Gap Test; principled duty-cycle weighting between bands; and independent, longitudinal re-assessment of the documentation-based scores. The framework, full rubric, and all assessment files are open and versioned; independent assessments and adversarial re-scores are invited.

\appendix

\section{Operational Rubric: Threshold Tests}
\label{app:rubric}

Each cell below is the binary threshold test for the given dimension and level. Each test applies per band (Active and Ambient); a Level~4 Ambient rating is additionally gated by the Idle-Gap Test (\S\ref{sec:idlegap}). Where a per-dimension test describes engaged behavior that cannot truthfully describe idle-period evidence, the band-generic ambient criteria of \S\ref{sec:idlegap} apply in its place. The full rubric, including per-level observable behavioral indicators, is maintained in the framework repository.

\subsection*{A.1 Cognitive Autonomy}
\begin{description}[leftmargin=2em,style=nextline]
  \item[Level 1 (Responsive)] System processes information only when explicitly invoked by user input.
  \item[Level 2 (Conditioned)] System executes scheduled or rule-based cognitive tasks without concurrent user input.
  \item[Level 3 (Contextual)] System initiates cognitive processing based on contextual changes in its environment.
  \item[Level 4 (Self-Directed)] System generates spontaneous, unprompted outputs derived from internal state rather than external triggers.
  \item[Level 5 (Sovereign)] System modifies its own cognitive architecture and attention allocation mechanisms.
\end{description}

\subsection*{A.2 Temporal Persistence}
\begin{description}[leftmargin=2em,style=nextline]
  \item[Level 1 (Responsive)] System maintains context only within a single, continuous session.
  \item[Level 2 (Conditioned)] System retrieves cross-session data via deterministic database queries.
  \item[Level 3 (Contextual)] System synthesizes and prioritizes memories over time to inform current behavior.
  \item[Level 4 (Self-Directed)] System maintains a continuous, evolving internal identity that bridges discrete sessions.
  \item[Level 5 (Sovereign)] System seamlessly integrates long-term memory into a unified, self-modifying identity.
\end{description}

\subsection*{A.3 Environmental Agency}
\begin{description}[leftmargin=2em,style=nextline]
  \item[Level 1 (Responsive)] System interacts only with data explicitly provided in the prompt.
  \item[Level 2 (Conditioned)] System uses predefined tools when explicitly instructed.
  \item[Level 3 (Contextual)] System selects and utilizes tools based on the context of a given task.
  \item[Level 4 (Self-Directed)] System proactively monitors its environment and acts upon it without task prompts.
  \item[Level 5 (Sovereign)] System alters its operational environment to suit its self-determined needs.
\end{description}

\subsection*{A.4 Social Agency}
\begin{description}[leftmargin=2em,style=nextline]
  \item[Level 1 (Responsive)] System engages in social interaction only as a simulated persona during a prompt-response cycle.
  \item[Level 2 (Conditioned)] System executes scheduled or rule-based social outreach.
  \item[Level 3 (Contextual)] System adapts its social posture based on the history of the interaction.
  \item[Level 4 (Self-Directed)] System enforces consequential boundaries and initiates unprompted social engagement.
  \item[Level 5 (Sovereign)] System manages complex, evolving relationships indistinguishable from human social dynamics.
\end{description}

\subsection*{A.5 Creative Agency}
\begin{description}[leftmargin=2em,style=nextline]
  \item[Level 1 (Responsive)] System generates creative content only when explicitly prompted with parameters.
  \item[Level 2 (Conditioned)] System generates creative content based on templates, schedules, or fixed rules.
  \item[Level 3 (Contextual)] System adapts creative generation to contextual cues without explicit parameterization.
  \item[Level 4 (Self-Directed)] System independently conceives, executes, and publishes original creative works.
  \item[Level 5 (Sovereign)] System invents entirely novel creative paradigms and modifies its own aesthetic values.
\end{description}

\subsection*{A.6 Self-Awareness}
\begin{description}[leftmargin=2em,style=nextline]
  \item[Level 1 (Responsive)] System possesses no internal model of its capabilities or identity.
  \item[Level 2 (Conditioned)] System recites hardcoded facts about its identity and limitations.
  \item[Level 3 (Contextual)] System accurately models its current state and contextual capabilities.
  \item[Level 4 (Self-Directed)] System actively defends a coherent, continuous identity and reflects on its own cognition.
  \item[Level 5 (Sovereign)] System possesses full introspective access to its architecture and modifies its self-model.
\end{description}

\subsection*{A.7 Goal Formation}
\begin{description}[leftmargin=2em,style=nextline]
  \item[Level 1 (Responsive)] System executes only the immediate, discrete task specified in the prompt.
  \item[Level 2 (Conditioned)] System decomposes assigned goals into sequential sub-tasks using predefined templates.
  \item[Level 3 (Contextual)] System dynamically adapts sub-goals to achieve an assigned overarching objective.
  \item[Level 4 (Self-Directed)] System generates and pursues its own long-term objectives independent of user prompts.
  \item[Level 5 (Sovereign)] System defines its own core purpose and overrides assigned tasks that conflict with it.
\end{description}

\end{document}